\documentclass[journal]{IEEEtai}

\usepackage{cite} 
\usepackage{color,array}
\usepackage{graphicx}
\usepackage{amsmath}
\usepackage{amssymb}
\usepackage{booktabs} 
\usepackage{multirow}
\usepackage{algorithm}
\usepackage{algorithmic}
\usepackage{xcolor}
\usepackage{colortbl} 
\usepackage{soul} 
\usepackage[colorlinks,urlcolor=blue,linkcolor=blue,citecolor=blue]{hyperref}

\begin{document}

\title{C3Net: Context-Contrast Network for Camouflaged Object Detection}

\author{Baber~Jan, Aiman~H.~El-Maleh, Abdul~Jabbar~Siddiqui, Abdul Bais and Saeed~Anwar
\thanks{Manuscript submitted to IEEE Transactions on Artificial Intelligence, Month Day, Year.}
\thanks{B. Jan, A. H. El-Maleh, and A. J. Siddiqui are with the Computer Engineering Department, King Fahd University of Petroleum and Minerals, Dhahran 31261, Saudi Arabia (e-mail: baberjan008@gmail.com; aimane@kfupm.edu.sa; abduljabbar.siddiqui@kfupm.edu.sa).}
\thanks{A. Bais is with Electronic Systems Engineering, University of Regina, Canada (e-mail: Abdul.Bais@uregina.ca).}
\thanks{S. Anwar is with the Department of Computer Science and Software Engineering, The University of Western Australia, Perth, WA 6009, Australia (e-mail: saeed.anwar@uwa.edu.au).}
\thanks{Corresponding author: Saeed Anwar}}

\markboth{IEEE TRANSACTIONS ON ARTIFICIAL INTELLIGENCE, VOL.~XX, NO.~X, MONTH~2025}%
{Jan \MakeLowercase{\textit{et al.}}: C3Net: Context-Contrast Network for Camouflaged Object Detection}

\maketitle

\begin{abstract}
Camouflaged object detection identifies objects that blend seamlessly with their surroundings through similar colors, textures, and patterns. This task challenges both traditional segmentation methods and modern foundation models, which fail dramatically on camouflaged objects. We identify six fundamental challenges in COD: Intrinsic Similarity, Edge Disruption, Extreme Scale Variation, Environmental Complexities, Contextual Dependencies, and Salient-Camouflaged Object Disambiguation. These challenges frequently co-occur and compound the difficulty of detection, requiring comprehensive architectural solutions. We propose C3Net, which addresses all challenges through a specialized dual-pathway decoder architecture. The Edge Refinement Pathway employs gradient-initialized Edge Enhancement Modules to recover precise boundaries from early features. The Contextual Localization Pathway utilizes our novel Image-based Context Guidance mechanism to achieve intrinsic saliency suppression without external models. An Attentive Fusion Module synergistically combines the two pathways via spatial gating. C3Net achieves state-of-the-art performance with S-measures of 0.898 on COD10K, 0.904 on CAMO, and 0.913 on NC4K, while maintaining efficient processing. C3Net demonstrates that complex, multifaceted detection challenges require architectural innovation, with specialized components working synergistically to achieve comprehensive coverage beyond isolated improvements. Code, model weights, and results are available at \url{https://github.com/Baber-Jan/C3Net}.
\end{abstract}

\begin{IEEEImpStatement}
C3Net advances camouflaged object detection with implications across multiple domains. Politically, standardized detection algorithms promote transparent and accountable automated systems. Economically, industrial defect detection applications can improve manufacturing quality control efficiency. Socially, enhanced polyp detection in colonoscopy screening supports early cancer diagnosis, addressing healthcare challenges where camouflaged lesions are frequently missed. Technologically, C3Net's dual-pathway architecture and intrinsic saliency suppression mechanism advance computer vision capabilities, demonstrating how architectural innovation solves complex detection challenges. Environmentally, automated wildlife monitoring enables efficient population tracking for conservation research. Legally, our research uses publicly available datasets without personal data collection, demonstrating privacy-conscious development practices. C3Net shows that specialized architectures can address multi-faceted vision challenges while maintaining ethical research standards. We encourage responsible deployment guided by domain-specific requirements and regulatory frameworks.
\end{IEEEImpStatement}


\section{Introduction}
\label{sec:introduction}

\IEEEPARstart{C}{amouflaged} objects blend into their surroundings through shared colors, textures, and patterns. Detecting such objects is termed Camouflaged Object Detection (COD) and differs fundamentally from conventional object detection~\cite{fan2021concealed}. While conventional detection targets objects with distinct visual features, camouflaged objects exhibit a high degree of intrinsic similarity to their backgrounds. This similarity renders standard segmentation methods ineffective, necessitating specialized approaches. State-of-the-art foundation models like SAM2~\cite{ravi2024sam2} achieve remarkable performance on general segmentation but experience dramatic degradation on camouflaged objects~\cite{tang2024evaluatingsam2srolecamouflaged}. This performance gap motivates the development of specialized COD architectures. COD enables critical applications including medical polyp detection~\cite{fan2020pranet}, wildlife monitoring~\cite{Wildlife}, and industrial inspection~\cite{bhajantri2006camouflage}.

\begin{figure}[t!]
    \centering
    \includegraphics[width=0.48\textwidth]{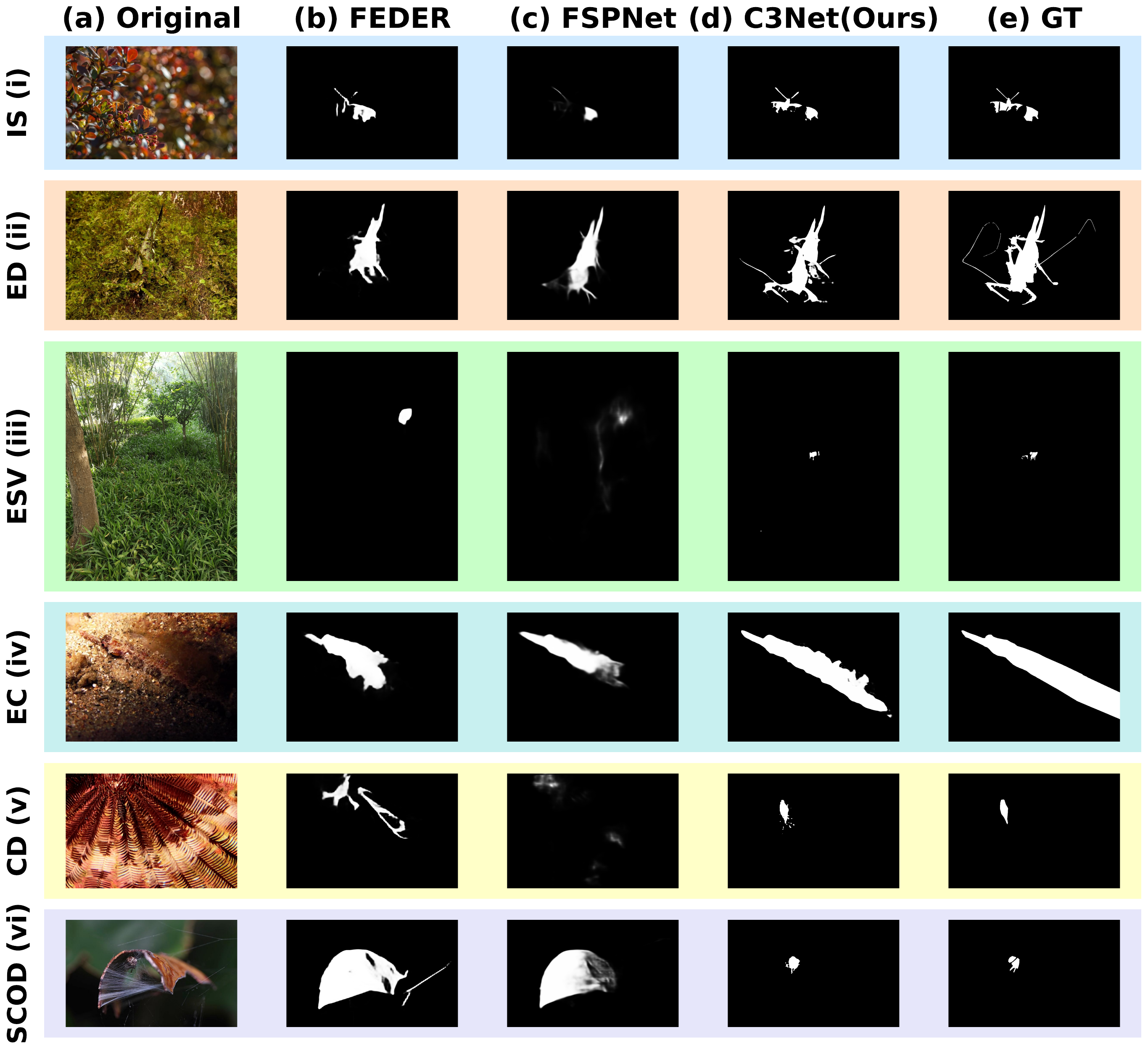}
    \caption{Visual comparison demonstrating C3Net's comprehensive handling of COD challenges. Each row illustrates a fundamental challenge: (i) Intrinsic Similarity (IS) - a camouflaged insect blends with leaves; (ii) Edge Disruption (ED) - an insect exhibits fragmented boundaries against ground; (iii) Extreme Scale Variation (ESV) - a bear barely visible in dense vegetation occupies minimal pixels; (iv) Environmental Complexities (EC) - shadows and terrain variations obscure half of the ground creature; (v) Contextual Dependencies (CD) - the insect requires global context for accurate segmentation; (vi) Salient-Camouflaged Object Disambiguation (SCOD) - a camouflaged insect must be distinguished from the prominent bark. C3Net consistently outperforms FEDER~\cite{He_2023_CVPR} (CNN-based SOTA) and FSPNet~\cite{huang2023feature} (ViT-based SOTA) across all challenges.}
    \label{fig:challenge_examples}
\end{figure}

Camouflaged object detection confronts six fundamental challenges that frequently co-occur and compound difficulty. Intrinsic Similarity (IS) forms the foundation where objects share identical colors, textures, and patterns with their backgrounds and thus become visually indistinguishable (Figure~\ref{fig:challenge_examples}, row i). This similarity directly contributes to Edge Disruption (ED), where object boundaries fragment or vanish entirely and make precise segmentation impossible (Figure~\ref{fig:challenge_examples}, row ii). These boundary ambiguities become critical with Extreme Scale Variation (ESV), where objects occupy minimal pixels or exhibit extreme aspect ratios and challenge detection networks (Figure~\ref{fig:challenge_examples}, row iii). Environmental Complexities (EC) further degrade visibility through shadows, occlusions, and changes in illumination that obscure already-weak boundaries (Figure~\ref{fig:challenge_examples}, row iv). Because local features fail under these conditions, Contextual Dependencies (CD) require models to integrate global scene information, yet local details remain essential for accuracy (Figure~\ref{fig:challenge_examples}, row v). Compounding these difficulties, Salient-Camouflaged Object Disambiguation (SCOD) occurs where scenes contain both camouflaged and salient objects, and models must detect only the camouflaged target while classifying the prominent salient object as background (Figure~\ref{fig:challenge_examples}, row vi). These interconnected challenges demand comprehensive architectures that address their complex interactions rather than isolated mechanisms.

Researchers have pursued various approaches to address these interconnected challenges, yet none achieve comprehensive coverage. Early CNN-based methods, such as SINet~\cite{Fan_2020_CVPR}, introduced specialized modules for extracting subtle cues but struggled with integrating global context. Multi-scale fusion~\cite{pang2022zoom} and iterative refinement~\cite{Jia_2022_CVPR} improved local feature processing while boundary precision remained limited. Transformer architectures~\cite{Yang_2021_ICCV,huang2023feature} achieved superior global reasoning but generated coarse segmentation masks that missed fine details. Joint salient and camouflaged object detection~\cite{li2021uncertainty} attempted to handle SCOD through multi-task learning, yet required external saliency models. These evolutionary improvements address specific challenges while leaving others unsolved. CNN methods excel at capturing local details but struggle with CD. Transformers capture global context but struggle with ED and ESV. Current architectures lack mechanisms to handle all six challenges systematically and particularly fail to suppress intrinsic saliency without external dependencies.

We propose C3Net (Context-Contrast Camouflaged Object Detection Network) to provide systematic coverage of all challenges. Our C3Net introduces specialized decoder pathways that process complementary visual cues rather than pursuing incremental improvements. Our approach separates edge refinement from contextual understanding, preventing signal dilution while ensuring synergistic operation. The architecture incorporates intrinsic mechanisms for each major challenge and integrates them through adaptive fusion. This design enables state-of-the-art performance across all benchmarks while maintaining processing efficiency. C3Net demonstrates that effective detection emerges from architectural integration rather than isolated components.

The main contributions of this paper are summarized as follows:
\begin{itemize}
    \item We design a dual-pathway decoder in which edge refinement and contextual localization operate at different feature levels. This separation prevents signal dilution and enables specialized learning for each visual cue type.

    \item We introduce Edge Enhancement Modules with multi-path convolutions initialized from gradient and Laplacian operators. These modules maintain classical edge detection principles while adapting to camouflage patterns through learning.

    \item We develop the Image-based Context Guidance mechanism that analyzes input appearance directly. This approach achieves intrinsic saliency suppression through contrast computation, eliminating the need for external models.

    \item We create an Attentive Fusion Module that spatially gates edge information via contextual pathways. This design emphasizes relevant boundaries while suppressing distractors at each spatial location.

    \item We formulate pathway-specific loss objectives with precision focus for saliency suppression and recall focus for complete capture. This strategy ensures each component learns its specialized role effectively.
\end{itemize}

\section{Related Work}
\label{sec:related_work}

Camouflaged object detection methods have evolved through distinct methodological approaches over the past decade. Each approach targets specific detection challenges while advancing the field toward more comprehensive solutions. This section examines these methodological directions and their contributions to addressing COD challenges.

\vspace{1mm}\noindent\textbf{Progressive Detection Strategies.} Early COD research established fundamental detection paradigms through progressive refinement. SINet~\cite{Fan_2020_CVPR} pioneered COD as a distinct task and introduced search-identification modules to handle high intrinsic similarity between objects and backgrounds. PFNet~\cite{Mei_2021_CVPR} advanced this approach by incorporating distraction mining to explicitly address false positives and false negatives through positioning and focus modules. These methods demonstrated success in identifying potential camouflaged regions and progressively refining predictions. However, these methods primarily focused on local-region analysis and progressive refinement, without comprehensive global scene understanding. This limitation motivated the exploration of architectures with enhanced global modeling capabilities.

\vspace{1mm}\noindent\textbf{Multi-Scale and Iterative Approaches.} Researchers developed multi-scale strategies to address the scale variation challenges in COD. ZoomNet~\cite{pang2022zoom} employed mixed-scale triplet networks to capture discriminative features at different zoom levels, while SegMaR~\cite{Jia_2022_CVPR} used iterative segment-magnify-reiterate strategies for progressive refinement. These methods improved object detection across scales through hierarchical processing. Yet, these methods primarily addressed scale-related aspects without comprehensive mechanisms for other challenges, such as saliency suppression and contextual dependencies. The approaches revealed that scale handling alone cannot substitute for a comprehensive understanding of features.

\vspace{1mm}\noindent\textbf{Transformer-Based Global Modeling.} The adoption of transformer architectures brought enhanced global context modeling to COD. FSPNet~\cite{huang2023feature} addressed the limitations of standard transformers in locality modeling through non-local token enhancement and feature shrinkage pyramids. CamoFormer~\cite{yin2024camoformer} employed masked separable attention to model foreground, background, and global context with distinct attention heads. These approaches significantly improved handling of contextual dependencies and global feature relationships. However, transformer architectures inherently prioritize global over local modeling, generating coarse feature representations. This architectural characteristic causes them to struggle with edge disruption (ED) and extreme scale variation (ESV), both of which require fine details. The trade-off reveals that global and local understanding require fundamentally different architectural treatments.

\vspace{1mm}\noindent\textbf{Edge and Boundary Enhancement.} Specialized methods emerged to address the boundary precision challenges in COD. BGNet~\cite{sun2022bgnet} combined boundary guidance with dual-branch global-local context integration to improve edge detection while maintaining contextual understanding. FEDER~\cite{He_2023_CVPR} explicitly tackled both intrinsic similarity and ambiguous boundaries through feature decomposition and ODE-inspired edge reconstruction. These approaches achieved significant improvements in boundary quality and demonstrated that edge enhancement benefits from integration with semantic understanding. However, their primary focus on boundary-related challenges left other aspects, such as extreme scale variation and comprehensive saliency handling, as secondary considerations. This specialization pattern shows that targeting specific challenges often comes at the expense of comprehensive coverage.

\vspace{1mm}\noindent\textbf{Joint Learning and Saliency Handling.} The contradictory relationship between salient and camouflaged objects motivated joint learning approaches. Salient-Camouflaged Object Disambiguation (SCOD) occurs when scenes contain both prominent salient objects and subtle camouflaged objects, where models must detect only the camouflaged target while classifying the salient object as background. UJSC~\cite{li2021uncertainty} pioneered this direction through uncertainty-aware training with data-wise correlation modeling, task-wise correlation modeling, and adversarial learning to distinguish between opposing visual characteristics. Zhao~et~al.~\cite{zhao2023nowhere} proposed saliency attribute transfer to spot camouflaged objects, while recent USCOD~\cite{zhou2024unconstrained} introduced the CS12K dataset with four scene types and the Camouflage-Saliency Confusion Score metric for comprehensive evaluation. These approaches demonstrated that explicitly modeling the SOD-COD relationship improved detection accuracy and provided systematic evaluation frameworks. However, standard COD benchmarks lack salient object annotations and thus models requiring saliency supervision cannot be fairly evaluated on established datasets. Joint training frameworks also introduced computational overhead and external dependencies through multi-task setups or saliency models. This reveals that while saliency handling is crucial for SCOD, intrinsic mechanisms without external dependencies remain an open challenge.

\vspace{1mm}\noindent\textbf{Integration and Hybrid Approaches.} Researchers combined these methodological directions to create comprehensive solutions. UJSC~\cite{li2021uncertainty} integrated joint saliency-camouflage learning with uncertainty modeling, while BGNet~\cite{sun2022bgnet} merged boundary guidance with global-local processing. HDPNet~\cite{he2025hdpnet} combined hourglass transformer structures with dual-path pyramids to balance global understanding and local detail preservation. These integration efforts demonstrated that combining complementary approaches could simultaneously address multiple challenges and achieve notable improvements across various benchmarks. However, the integration strategies primarily relied on established fusion techniques that may not fully exploit the complementary nature of different components. The field offers opportunities for architectures in which specialized components interact through learned, adaptive mechanisms that enhance their collective effectiveness. This suggests potential for designs that achieve integration through architectural innovation rather than additive complexity.

\vspace{1mm}\noindent\textbf{Existing Gaps and Opportunities.} An analysis of existing methods reveals systematic gaps in comprehensively addressing COD challenges. While individual approaches excel at specific aspects, no current architecture provides integrated solutions for all six fundamental challenges. Edge-focused methods achieve precise boundaries but lack contextual understanding for SCOD. Transformer approaches capture global dependencies but sacrifice fine details essential for ED and ESV. Joint learning frameworks handle saliency disambiguation but introduce external dependencies that complicate deployment. Most critically, existing architectures lack intrinsic mechanisms for saliency suppression and rely on external models or multi-task setups. The field requires architectures that address these challenges through synergistic design rather than isolated improvements. Opportunities exist for methods that combine specialized processing pathways with adaptive fusion mechanisms, thereby maintaining both global coherence and local precision. Such architectures could achieve comprehensive coverage through architectural innovation rather than component accumulation.
\begin{figure*}[t!]
    \centering
    \includegraphics[width=0.95\textwidth]{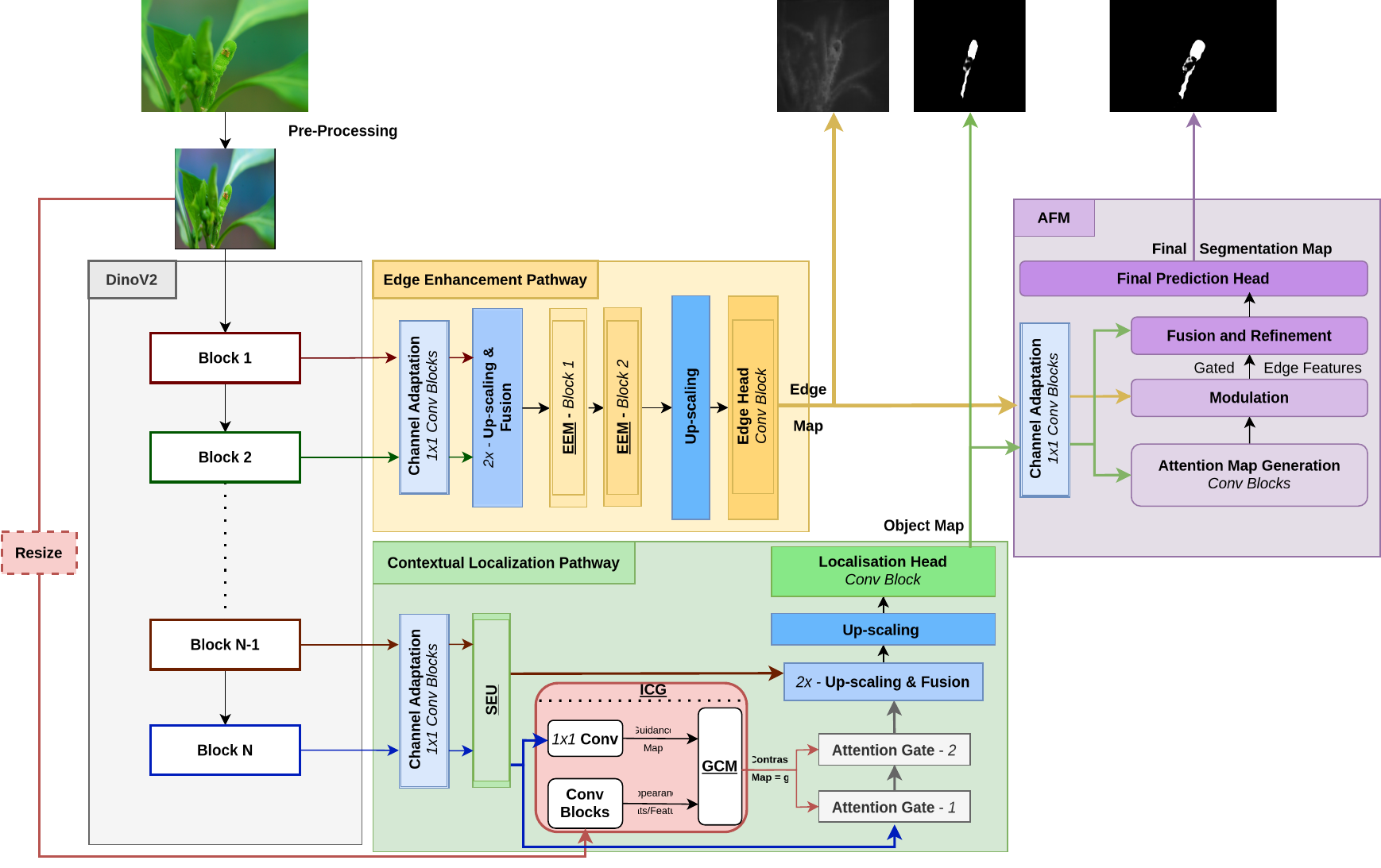}
    \caption{Overview of the C3Net architecture. Input images are preprocessed and encoded to extract multi-scale features. Our dual-pathway decoder comprises two specialized branches. The Edge Refinement Pathway (top) processes early features through Edge Enhancement Modules (EEMs) to produce detailed edge maps. The Contextual Localization Pathway (bottom) processes deep features through Semantic Enhancement Units (SEUs) and our Image-based Context Guidance (ICG) mechanism to generate object maps with suppressed saliency. The ICG contains three components: appearance analysis from the input image, the Guided Contrast Module (GCM) for foreground-background differentiation, and iterative attention gates for saliency suppression. The Attentive Fusion Module (AFM) combines the outputs of both pathways through spatial gating to produce the final segmentation. Deep supervision is applied at three points: edge map, object map, and final prediction.}
    \label{fig:c3net_architecture}
\end{figure*}

\section{Methodology}
\label{sec:methodology}


Our proposed C3Net model follows an encoder-decoder architecture with specialized components, as shown in Figure~\ref{fig:c3net_architecture}. Here, we provide an overview of our C3Net. Initially, a vision transformer encoder extracts features from the input image $\mathbf{I}$ containing camouflaged objects, which are then fed into our dual-pathway decoder architecture. Our Edge Enhancement Modules process early features to recover precise boundaries. Similarly, our Image-based Context Guidance mechanism processes deep features to identify objects while suppressing salient distractors. Both mentioned modules have different pathways that utilize content-adaptive upsampling to preserve fine details. Moreover, our Attentive Fusion Module combines complementary outputs, and finally, multiple losses provide supervision at three prediction points.



C3Net learns a function $f: \mathbf{I} \rightarrow \mathbf{M}$ to predict a binary segmentation mask ($\mathbf{M} \in \{0,1\}^{B \times H \times W}$) where 1 indicates camouflaged object pixels. The following subsections detail our dual-pathway architecture and its components.

\subsection{Feature Encoding}
\label{ssec:feature_encoding}

We extract features from input images $\mathbf{I}$ using DINOv2 with register tokens~\cite{darcet2024visiontransformersneedregisters} as the backbone encoder. DINOv2 is a vision transformer pre-trained via self-supervision without semantic labels. It learns unbiased features for detecting atypical objects, with register tokens that prevent artifacts and ensure clean, accurate upsampling for dense prediction.
The encoder contains $N$ transformer blocks that generate features at consistent spatial resolution. For the Large variant, $N=24$ blocks produce features with dimensions $\mathbf{F}_i$.

We select four specific feature sets based on their complementary properties, which serve as inputs to our dual-pathway decoder architecture. Early features $\mathbf{F}_1$ and $\mathbf{F}_2$ from the first two blocks retain fine-grained spatial details and high-frequency information essential for edge detection. Deep features $\mathbf{F}_{n-1}$ and $\mathbf{F}_n$ from the final two blocks capture abstract semantic representations necessary for object understanding and contextual reasoning. This selection strikes a balance between spatial precision and semantic richness. 

\subsection{Dual-Pathway Decoder}
\label{ssec:dual_pathway_decoder}
Our decoder separates edge and contextual processing into different pathways. This design arises from the observation that boundary detection and semantic understanding require fundamentally different feature-processing strategies. Unified decoders often dilute these distinct features through shared transformations. Our dual-pathway approach prevents this interference while enabling targeted learning. The Edge Refinement Pathway processes early encoder features $\mathbf{F}_1$ and $\mathbf{F}_2$, which are rich in spatial details. The Contextual Localization Pathway processes deep encoder features $\mathbf{F}_{n-1}$ and $\mathbf{F}_n$, which contain semantic information. Both pathways operate at progressively increasing resolutions through content-adaptive upsampling. Their outputs merge through learned spatial gating rather than simple concatenation. This architectural separation enables each pathway to develop specialized representations optimized for specific tasks.

\subsubsection{Edge Refinement Pathway (ERP)}
\label{sssec:edge_refinement_pathway}

The ERP extracts precise object boundaries from early encoder features $\mathbf{F}_1$ and $\mathbf{F}_2$. 
These features are first reduced in channel dimension via $1 \times 1$ convolutions, optimizing the representations for edge detection.
Then, these features are progressively upsampled using DySample~\cite{liu2023learning} module, which dynamically learns sampling locations based on feature content, thereby preserving fine boundary details during resolution recovery. After upsampling, these features are concatenated, followed by a convolution to produce $\mathbf{F}_{fused}^{edge}$, which captures complementary edge cues at different encoding depths. The fused representation is then processed through two cascaded Edge Enhancement Modules.

\textit{Edge Enhancement Module (EEM).} is our core component designed to amplify edge-specific signals in camouflaged object detection. Each module employs a multi-path convolutional architecture that processes features across three parallel branches. The design leverages classical edge detection principles within a learnable framework. This combination provides strong inductive bias while maintaining adaptability to camouflage-specific patterns. The three paths within each EEM serve distinct purposes: i) Context Path: Captures general spatial patterns through depthwise separable convolutions, extracting contextual edge information without directional bias. ii)  Gradient Path: Detects directional changes using learnable group convolutions initialized with Sobel weights. Sobel operators excel at detecting gradual intensity transitions—precisely the subtle boundaries found in camouflaged objects where sharp edges are deliberately avoided. iii) Discontinuity Path: Identifies edge points through learnable group convolutions initialized with Laplacian weights. The Laplacian's zero-crossing detection captures texture boundaries where camouflaged patterns meet backgrounds, complementing gradient-based detection.

Each EEM processes the input features $\mathbf{F}_{edge}^{in}$ through the three paths simultaneously. Outputs from all three paths are concatenated along the channel dimension to combine their complementary edge information. This representation undergoes channel-wise recalibration using ECA~\cite{wang2020eca}. 
The module produces output features $\mathbf{F}_{edge}^{out}$ that contain amplified edge signals. The ERP employs two EEMs in sequence to progressively enhance edge quality. The fused features $\mathbf{F}_{fused}^{edge}$ are fed into the first EEM for initial edge extraction. This module captures coarse edge structures and produces refined features. These refined features then enter the second EEM for further enhancement. The second EEM extracts fine boundary details and outputs the final edge features $\mathbf{F}_{edge}$. Additionally, these features are fed into a prediction head to generate an edge map $\mathbf{P}_{edge}$. Thus, the pathway generates two outputs: $\mathbf{F}_{edge}$ for subsequent fusion and $\mathbf{P}_{edge}$ for supervision. 

\subsubsection{Contextual Localization Pathway (CLP)}
\label{sssec:contextual_localization_pathway}

The CLP identifies semantic regions of camouflaged objects while suppressing salient distractors. This pathway receives encoder features $\mathbf{F}_{n-1}$ and $\mathbf{F}_n$ as inputs, which contain rich semantic information. Each feature is then processed through a Semantic Enhancement Unit (SEU).


\textit{Semantic Enhancement Unit (SEU).} is our component that refines semantic features for improved object localization. Each unit processes its input adapted features through a series of transformations. First, depthwise separable convolutions extract initial spatial patterns $\mathbf{F}_{loc}^{conv}$ while maintaining efficiency. 
These then undergo enhancement through spatial and channel attention mechanisms applied in parallel: i) Spatial Attention: Identifies discriminative regions by learning where to focus, adapting the concatenation strategy from CBAM~\cite{Woo_2018_ECCV}. The feature maps are aggregated using both average and max pooling along the channel dimension, capturing different spatial statistics, and are concatenated to form a comprehensive spatial descriptor. The descriptor passes through convolutional layers to generate spatial attention weights. ii) Channel Attention: Determines feature importance by recalibrating channel-wise responses~\cite{wang2020eca}. It learns which channels contain the most discriminative information for camouflaged objects.

Both attention outputs are combined with the convolutional features $\mathbf{F}_{loc}^{conv}$ through element-wise multiplication. This combination enhances discriminative semantic patterns while suppressing irrelevant information. Finally, a residual connection adds the SEU's input-adapted features to these attention-enhanced outputs. Thus, providing refined features that emphasize both spatially and channel-wise important information for contextual understanding.

As discussed above, each enhanced feature has distinct roles: i) Penultimate Layer Processing: The first SEU enhances the adapted $\mathbf{F}_{n-1}$ to produce enhanced features $\mathbf{F}_{n-1}^{enh}$, which are then upsampled using DySample~\cite{liu2023learning} to produce $\mathbf{F}_{n-1}^{up}$, recovering spatial resolution while preserving localization details for fusion. ii) Final Layer Processing: The second SEU enhances the adapted $\mathbf{F}_n$ to produce $\mathbf{F}_n^{enh}$. These features are fed directly into our Image-based Context Guidance mechanism for saliency suppression and context-aware modulation. This design leverages deep features for semantic context and shallow features for spatial precision, optimizing each for effective camouflaged object detection.


\textit{Image-based Context Guidance (ICG).} is for intrinsic saliency suppression and processes $\mathbf{F}_n^{enh}$ from the final encoder layer and the input image. This mechanism integrates three components to suppress salient distractors while enhancing camouflaged object cues: i) Appearance Analysis: Extracts low-level visual patterns directly from the input image. A lightweight CNN consisting of two convolutional blocks processes $\mathbf{I}$ to produce appearance features $\mathbf{A_f}$ that capture colour distributions and texture patterns independent of learned semantic representations and provide essential visual cues for contrast computation. ii) Guided Contrast Computation: Generates spatially-aware contrast maps using the appearance features $\mathbf{A_f}$. First, an initial object hypothesis $\mathbf{G}$ is generated from $\mathbf{F}_n^{enh}$. Then, the Guided Contrast Module (GCM) uses hypothesis $\mathbf{G}$ to weigh $\mathbf{A_f}$ spatially. This aggregation extracts local foreground representations using $\mathbf{G}$ as continuous weights. Meanwhile, global pooling on confident background regions produces robust background representations. Specifically, the GCM computes:
\begin{align}
    \mathbf{f}_{fg} &= \frac{\sum_{(i,j)} \mathbf{A_f}_{:,i,j} \cdot \mathbf{G}_{i,j}}{\sum_{(i,j)} \mathbf{G}_{i,j}}, \\
    \mathbf{f}_{bg} &= \frac{1}{|\Omega_{bg}|} \sum_{(i,j) \in \Omega_{bg}} \mathbf{A_f}_{:,i,j},
\end{align}
where $\Omega_{bg} = \{(i,j) | \mathbf{G}_{i,j} < 0.1\}$. The contrast map is generated by a contrast computation network that processes appearance features and extracted representations. This contrast highlights regions that subtly differ from their surroundings. 
ii) Iterative Attention Gating: Progressively refines features through dual attention gates. The first attention gate uses a contrast map as a gating signal to modulate $\mathbf{F}_n^{enh}$, producing intermediate features $\mathbf{F}_{loc}^{(1)}$. Subsequently, the second attention gate further refines intermediate features $\mathbf{F}_{loc}^{(1)}$ using the same contrast map, yielding final modulated features $\mathbf{F}_{loc}^{mod}$. This iterative process suppresses regions with high saliency but low contrast while enhancing regions with subtle differences. The dual-gate design ensures robust distractor suppression without losing fine camouflage cues. The ICG achieves intrinsic saliency suppression by combining appearance analysis and semantic features, enabling camouflaged object detection without external saliency models.


Furthermore, the ICG's output undergoes multi-scale fusion to produce the final contextual features. The modulated features $\mathbf{F}_{loc}^{mod}$ are first upsampled~\cite{liu2023learning} to produce upsampled modulated features $\mathbf{F}_{loc}^{up}$ for accurate localization. These features are then fused with upsampled features $\mathbf{F}_{n-1}^{up}$. The fusion combines the context-aware deep features with the spatially detailed shallow features. This combination yields fused features $\mathbf{F}_{fused}^{loc}$.

The fused features undergo three processing stages to generate the final contextual features. First, a lightweight refinement module consisting of depthwise separable convolutions smooths the combined features and resolves potential conflicts between deep and shallow representations. Next, an initial upsampling recovers spatial detail for precise localization, and the subsequent upsampling produces the contextual localization features $\mathbf{F}_{loc}$. Additionally, a prediction head generates an intermediate object map by applying a $1 \times 1$ convolution followed by a sigmoid activation to produce a prediction map $\mathbf{P}_{loc}$. Thus, the CLP outputs both localization features $\mathbf{F}_{loc}$ for final fusion and prediction map $\mathbf{P}_{loc}$ for loss computation.



\subsection{Attentive Fusion Module}
\label{ssec:attentive_fusion_module}

The Attentive Fusion Module combines the complementary outputs from both pathways to generate the final segmentation mask. This module receives edge features $\mathbf{F}_{edge}$ from the ERP and localization features $\mathbf{F}_{loc}$ from the CLP. The fusion process leverages spatial attention to weigh each pathway's contribution according to its spatial relevance. First, both feature sets are transformed into $\mathbf{F}_{edge}^{adapt}$ and $\mathbf{F}_{loc}^{adapt}$ via channel reduction, integrating effective information while maintaining pathway-specific characteristics.


Subsequently, a spatial attention mechanism computes attention weights from the adapted localization features $\mathbf{F}_{loc}^{adapt}$ to guide selective fusion. The attention generation applies global average and max pooling operations to extract comprehensive spatial statistics. These pooled representations are concatenated and processed by convolutional layers to generate a spatial attention map, $\mathbf{A}_{spatial}$, that identifies regions requiring enhanced boundary precision.

The spatial attention map $\mathbf{A}_{spatial}$ then modulates the adapted edge features $\mathbf{F}_{edge}^{adapt}$ through element-wise multiplication to emphasize boundary information selectively. This gating operation produces edge features $\mathbf{F}_{edge}^{att} = \mathbf{F}_{edge}^{adapt} \odot \mathbf{A}_{spatial}$ that focus on contextually relevant boundaries while suppressing spurious edge responses. Finally, the $\mathbf{F}_{edge}^{att}$ and $\mathbf{F}_{loc}^{adapt}$ are concatenated and refined through two convolutional blocks, producing a unified set of features, $\mathbf{F}_{fused}^{final}$.
The final prediction head generates a segmentation mask, $\mathbf{P}_{final}$, for accurate camouflaged object detection.

\subsection{Loss Function Strategy}
\label{ssec:loss_function}

The training objective combines three loss components corresponding to the edge prediction $\mathbf{P}_{edge}$, contextual prediction $\mathbf{P}_{loc}$, and final prediction $\mathbf{P}_{final}$. 
The total loss function is formulated as:
\begin{equation}
\mathcal{L}_{total} = w_{final} \mathcal{L}_{final} + w_{loc} \mathcal{L}_{loc} + w_{edge} \mathcal{L}_{edge},
\end{equation}
where $w_{final}$, $w_{loc}$, and $w_{edge}$ are weighting coefficients that balance the contribution of each component during training.

The edge loss $\mathcal{L}_{edge}$ supervises the ERP to enhance boundary detection capabilities. This loss combines spatially-weighted Focal Loss~\cite{lin2017focal} to address class imbalance around object boundaries and Total Variation (TV) loss to encourage smoothness in predicted background regions. 

The contextual loss $\mathcal{L}_{loc}$ guides the CLP to achieve accurate object localization with saliency suppression. This loss pairs Focal Loss with Tversky Loss~\cite{salehi2017tversky} configured to prioritize precision over recall. The precision emphasis reinforces the ICG mechanism's effectiveness in suppressing salient false positives, directly addressing the SCOD challenge through targeted optimization.

The final loss $\mathcal{L}_{final}$ optimizes the fused prediction for comprehensive object detection performance. This component combines Focal and Tversky losses configured to emphasize recall, thereby promoting complete object capture. Additionally, a weighted Dice Loss term, calculated specifically on ground-truth edge pixels, provides final refinement signals for boundary sharpness. This multi-metric approach ensures both accurate localization and precise boundary delineation.

Furthermore, instance-level weighting based on object size is applied to $\mathcal{L}_{loc}$ and $\mathcal{L}_{final}$ to address scale variation. Objects with smaller foreground ratios receive higher weights to prevent bias toward larger objects, thus improving detection performance.
\section{Experiments and Evaluation}
\label{sec:experiments}

\begin{table*}[t]
    \caption{Quantitative comparison with SOTA methods on benchmark datasets.
    Notes: $\uparrow$/$\downarrow$ denotes that larger/smaller is better.
    The best values are in \textbf{\color{red}bold red}, and the second best are \underline{\color{blue}{underlined in blue}}.
    Light-gray rows are CNN-based methods; white rows are Transformer-based methods. All results are obtained from author-provided predictions for fair comparison.}
    \centering
    \resizebox{\textwidth}{!}{
    \begin{tabular}{l|ccccc|ccccc|ccccc}
        \toprule
        \multirow{2}{*}{Methods}
        & \multicolumn{5}{c|}{CAMO (250)}
        & \multicolumn{5}{c|}{COD10K (2,026)}
        & \multicolumn{5}{c}{NC4K (4,121)}\\
        \cline{2-16}
        & $S_{\alpha}\uparrow$ & $F_\beta^w\uparrow$ & $F_\beta^m\uparrow$ & $E_\phi\uparrow$ & $\mathcal{M}\downarrow$
        & $S_{\alpha}\uparrow$ & $F_\beta^w\uparrow$ & $F_\beta^m\uparrow$ & $E_\phi\uparrow$ & $\mathcal{M}\downarrow$
        & $S_{\alpha}\uparrow$ & $F_\beta^w\uparrow$ & $F_\beta^m\uparrow$ & $E_\phi\uparrow$ & $\mathcal{M}\downarrow$\\
        \midrule
        \rowcolor{gray!15}
        SINet$_{20}$~\cite{Fan_2020_CVPR}
         & .751 & .606 & .675 & .831 & .100
         & .771 & .551 & .634 & .868 & .051
         & .808 & .723 & .769 & .883 & .058 \\
        \rowcolor{gray!15}
        SLSR$_{21}$~\cite{yunqiu_cod21}
         & .787 & .674 & .744 & .854 & .080
         & .804 & .673 & .715 & .892 & .037
         & .840 & .766 & .804 & .907 & .048 \\
        \rowcolor{gray!15}
        PFNet$_{21}$~\cite{Mei_2021_CVPR}
         & .782 & .695 & .746 & .855 & .085
         & .800 & .660 & .701 & .890 & .040
         & .829 & .745 & .784 & .898 & .053 \\
        \rowcolor{gray!15}
        MGL$_{21}$~\cite{zhai2021Mutual}
         & .775 & .673 & .726 & .842 & .088
         & .814 & .666 & .711 & .890 & .035
         & .833 & .740 & .782 & .893 & .052 \\
        \rowcolor{gray!15}
        UJSC$_{21}$~\cite{li2021uncertainty}
         & .800 & .728 & .772 & .873 & .073
         & .809 & .684 & .721 & .891 & .035
         & .842 & .771 & .806 & .907 & .047 \\
        \rowcolor{gray!15}
        C$^2$FNet$_{21}$~\cite{sun2021c2fnet}
         & .796 & .719 & .762 & .864 & .080
         & .813 & .686 & .723 & .900 & .036
         & .838 & .762 & .795 & .904 & .049 \\
        \rowcolor{gray!15}
        UGTR$_{21}$~\cite{Yang_2021_ICCV}
         & .784 & .684 & .735 & .851 & .086
         & .817 & .666 & .712 & .890 & .036
         & .839 & .747 & .787 & .899 & .052 \\
        \rowcolor{gray!15}
        PreyNet$_{22}$~\cite{10.1145/3503161.3548178}
         & .790 & .708 & .757 & .857 & .077
         & .813 & .697 & .736 & .891 & .034
         & - & - & - & - & - \\
        \rowcolor{gray!15}
        BSA-Net$_{22}$~\cite{Zhu_Li_Xie_Yan_Liang_Chen_Wei_Qin_2022}
         & .794 & .717 & .763 & .867 & .079
         & .818 & .699 & .738 & .901 & .034
         & .841 & .771 & .808 & .907 & .048 \\
        \rowcolor{gray!15}
        OCE-Net$_{22}$~\cite{9706783}
         & .802 & .723 & .766 & .865 & .080
         & .827 & .707 & .741 & .905 & .033
         & .853 & .785 & .818 & .913 & .045 \\
        \rowcolor{gray!15}
        BGNet$_{22}$~\cite{sun2022bgnet}
         & .812 & .749 & .789 & .882 & .073
         & .831 & .722 & .753 & .911 & .033
         & .851 & .788 & .820 & .916 & .044 \\
        \rowcolor{gray!15}
        SegMaR$_{22}$~\cite{Jia_2022_CVPR}
         & .815 & .795 & .794 & .884 & .071
         & .833 & .724 & .757 & .906 & .034
         & .841 & .781 & .820 & .907 & .046 \\
        \rowcolor{gray!15}
        ZoomNet$_{22}$~\cite{pang2022zoom}
         & .820 & .752 & .794 & .892 & .066
         & .830 & .729 & .766 & .911 & .029
         & .853 & .784 & .818 & .912 & .043 \\
        \rowcolor{gray!15}
        SINet-v2$_{22}$~\cite{fan2021concealed}
         & .820 & .743 & .782 & .895 & .070
         & .815 & .680 & .718 & .906 & .037
         & .847 & .770 & .805 & .914 & .048 \\
        \rowcolor{gray!15}
        FDNet$_{22}$~\cite{zhong2022detecting}
         & .828 & .748 & .781 & .883 & .068
         & .832 & .706 & .733 & .907 & .033
         & .834 & .750 & .784 & .893 & .051 \\
        \rowcolor{gray!15}
        DTINet$_{22}$~\cite{liu2022boosting}
         & .856 & .796 & - & .916 & .050
         & .824 & .695 & - & .896 & .034
         & .863 & .792 & - & .917 & .041 \\
        \rowcolor{gray!15}
        OSFormer$_{22}$~\cite{pei2022osformer}
         & .799 & - & - & .858 & .073
         & .811 & - & - & .881 & .034
         & .832 & - & - & .905 & .049 \\
        FSPNet$_{23}$~\cite{huang2023feature}
         & .856 & .799 & .830 & .928 & .050
         & .851 & .735 & .769 & .930 & .026
         & .879 & .816 & .843 & .937 & .035 \\
        TPRNet$_{22}$~\cite{zhang2023tprnet}
         & .814 & .781 & - & - & .076
         & .829 & .725 & - & - & .034
         & .854 & .790 & - & - & .047 \\
        FPNet$_{23}$~\cite{cong2023frequency}
         & .852 & .806 & - & .905 & .056
         & .850 & .748 & - & .913 & .029
         & - & - & - & - & - \\
        OPNet$_{23}$~\cite{mei2023camouflaged}
         & .858 & .817 & - & .915 & .050
         & .857 & .767 & - & .919 & .026
         & .883 & .838 & - & .932 & .034 \\
        HitNet$_{23}$~\cite{hu2023high}
         & .844 & .801 & - & .902 & .057
         & .868 & .798 & - & .932 & .024
         & .870 & .825 & - & .921 & .039 \\
        SAM-Auto$_{23}$~\cite{Kirillov_2023_ICCV,tang2023samsegmentanythingsam}
         & .684 & .606 & .680 & .687 & .132
         & .783 & .701 & .756 & .798 & .050
         & .767 & .696 & .752 & .776 & .078 \\
        SAM-Prompt$_{23}$~\cite{Kirillov_2023_ICCV,tang2023samsegmentanythingsam}
         & .647 & .520 & - & - & .141
         & .696 & .552 & - & - & .094
         & .699 & .591 & - & - & .115 \\
        \rowcolor{gray!15}
        FEDER$_{23}$~\cite{He_2023_CVPR}
         & .807 & .785 & \underline{\color{blue}{.873}} & \underline{\color{blue}{.947}} & .069
         & .823 & .740 & \textbf{\color{red}{.900}} & .911 & .032
         & .846 & .817 & \textbf{\color{red}{.905}} & .916 & .045 \\
        SAM2-Auto$_{24}$~\cite{ravi2024sam2,tang2024evaluatingsam2srolecamouflaged}
         & .444 & .184 & .207 & .401 & .236
         & .549 & .271 & .291 & .521 & .134
         & .512 & .251 & .268 & .482 & .186 \\
        SAM2-Prompt$_{24}$~\cite{ravi2024sam2,tang2024evaluatingsam2srolecamouflaged}
         & .722 & .633 & - & - & .114
         & .754 & .640 & - & - & .078
         & .776 & .700 & - & - & .085 \\
        FocusDiffuser$_{25}$~\cite{zhao2025focusdiffuser}
         & .881 & \underline{\color{blue}{.851}} & - & .939 & .042
         & \underline{\color{blue}{.875}} & \underline{\color{blue}{.809}} & - & \underline{\color{blue}{.939}} & \underline{\color{blue}{.020}}
         & .891 & \underline{\color{blue}{.854}} & - & .940 & \underline{\color{blue}{.029}} \\
        FSEL$_{25}$~\cite{sun2025frequency}
         & \underline{\color{blue}{.885}} & \underline{\color{blue}{.851}} & .864 & .942 & \underline{\color{blue}{.040}}
         & .873 & .800 & .796 & .928 & .021
         & \underline{\color{blue}{.892}} & .853 & .864 & \underline{\color{blue}{.941}} & .030 \\
         \midrule
        \textbf{C3Net (Ours)}
         & \textbf{\color{red}{.904}} & \textbf{\color{red}{.889}} & \textbf{\color{red}{.896}} & \textbf{\color{red}{.951}} & \textbf{\color{red}{.0311}}
         & \textbf{\color{red}{.898}} & \textbf{\color{red}{.851}} & \underline{\color{blue}{.859}} & \textbf{\color{red}{.961}} & \textbf{\color{red}{.0162}}
         & \textbf{\color{red}{.913}} & \textbf{\color{red}{.895}} & \underline{\color{blue}{.903}} & \textbf{\color{red}{.958}} & \textbf{\color{red}{.0220}} \\
        \bottomrule
    \end{tabular}}

\label{tab:comparison}
\vspace{-3mm}
\end{table*}

\subsection{Experimental Setup}
\label{subsec:exp_setup}

This section details the experimental configuration used to evaluate C3Net. We first describe the benchmark datasets employed for training and testing. We then explain the evaluation metrics used to assess performance. Finally, we present the implementation details, including training procedures and hyperparameter settings.

\paragraph{Datasets.} We evaluate our C3Net on three established COD benchmarks widely adopted in the literature. 
All datasets include pixel-level ground truth annotations for precise evaluation. Additionally, we generate edge maps using the Canny detector on ground truth masks with a 5-pixel width. These edge maps serve as supervision signals for our ERP during training. The first benchmark is \emph{COD10K}~\cite{Fan_2020_CVPR}, which contains 10,000 images, making it the largest COD dataset. Among these images, 5,066 depict camouflaged instances spanning 10 super-classes and 78 sub-classes. These classes cover aquatic, terrestrial, flying, and amphibious categories. We follow the standard split using 3,040 images for training and 2,026 for testing. The second benchmark is \emph{CAMO}~\cite{anabranch_aux_classification}, which provides 1,250 images featuring both natural and artificial camouflage. This dataset complements COD10K by including eight categories. 
We use the standard protocol of 1,000 images for training and 250 for testing. The third benchmark is \emph{NC4K}~\cite{yunqiu_cod21}, which serves as a test-only evaluation with 4,121 images. This dataset pushes detection limits by challenging natural camouflage in the presence of extremely similar backgrounds. It specifically tests cross-dataset generalization without fine-tuning. Following established practice, we train C3Net on the training sets of COD10K and CAMO to ensure a fair comparison. We then evaluate on all three benchmarks to assess both in-domain and cross-domain performance. NC4K serves as a held-out test set to measure cross-dataset generalization.

\paragraph{Evaluation Metrics.} We evaluate using five standard COD metrics. \emph{Structure Measure} ($S_\alpha$)~\cite{Cheng2021sMeasure} assesses structural similarity with $\alpha=0.5$. \emph{Enhanced Alignment Measure} ($E_\phi$)~\cite{eMeasure} combines pixel-level and global statistics. \emph{Weighted F-Measure} ($F_\beta^w$)~\cite{Margolin_2014_CVPR} and \emph{Mean F-Measure} ($F_\beta^m$)~\cite{meanFMeasure} evaluate precision-recall with $\beta^2=0.3$. \emph{Mean Absolute Error} ($\mathcal{M}$) measures pixel-wise differences where lower is better. These metrics comprehensively evaluate structural accuracy and boundary precision.

\paragraph{Implementation Details.} C3Net uses the PyTorch framework on NVIDIA H100 GPUs. Images are resized to $392 \times 392$ pixels and normalized with $\mu=[0.485, 0.456, 0.406]$ and $\sigma=[0.229, 0.224, 0.225]$. We use the pretrained DINOv2-Large model, extracting features from blocks [1, 2, 23, 24]. Decoder channels are [512, 256, 128] with an output channel of 128. The Edge pathway utilizes two enhancement blocks, whereas the contextual pathway employs one. ICG uses 32 appearance channels and 16 contrast channels. DySample uses the 'lp' style with four groups. The fusion head uses 128 channels with two refinement blocks. Loss weights are $w_{edge}=1.0$, $w_{loc}=1.15$, and $w_{final}=1.2$. Focal loss has the $\alpha=0.25$ and $\gamma=3.0$. Tversky for $\mathcal{L}_{loc}$ uses $\alpha=0.6$ and $\beta=0.4$ while $\mathcal{L}_{final}$ uses $\alpha=0.4$ and $\beta=0.6$. Edge loss employs a focal weight of 5.0 and a TV weight of 0.15, with a cutoff of 5. Final loss utilizes edge Dice weight 0.2. Instance weighting uses a factor of 3.0 with thresholds of 0.02 and 0.8. Training uses AdamW for 200 epochs with learning rates of $1 \times 10^{-4}$ for the decoder and $2 \times 10^{-5}$ for the encoder. Weight decay is 0.01, and gradient clipping is 1.0. We employ automatic mixed precision for efficiency. ReduceLROnPlateau uses a factor of 0.5, patience of 10, and a minimum of $1 \times 10^{-6}$. Batch size is 128. Inference uses identical preprocessing.

\subsection{Comparison with State-of-the-Art}
\label{subsec:sota_comparison}

We benchmark C3Net against a comprehensive suite of 29 recent COD methods, including prominent CNN-based models, transformer-based approaches, and foundation models. 

\paragraph{Quantitative Results.} Table~\ref{tab:comparison} demonstrates C3Net's state-of-the-art performance across benchmarks. On the primary COD10K benchmark, we achieve leading results with $S_\alpha$ of 0.898, $F_\beta^w$ of 0.851, $F_\beta^m$ of 0.859, and $E_\phi$ of 0.961. Our model outperforms the next-best method by 2.6\% in structure measure ($S_\alpha$) and 2.3\% in enhanced alignment measure ($E_\phi$), confirming that our dual-pathway design effectively captures both global coherence and local precision. The MAE of 0.0162 demonstrates precise pixel-level accuracy. This performance extends to CAMO, where we achieve $S_\alpha$ of 0.904, $F_\beta^w$ of 0.889, and the leading $\mathcal{M}$ of 0.0311, improving over the next-best by 2.1\% in structure measure. On NC4K, we maintain leadership with $S_\alpha$ of 0.913 and $E_\phi$ of 0.958. Across all three benchmarks, C3Net achieves the best performance on 13 of 15 metrics, validating our comprehensive challenge coverage rather than dataset-specific tuning. Foundation models confirm the necessity of specialized architecture, with SAM achieving only $S_\alpha=0.783$ and SAM2 catastrophically failing at $S_\alpha=0.549$. The performance gap between C3Net and SAM2 exceeds 63\% for the $S_\alpha$ metric, quantifying why general segmentation fails on camouflaged objects. These results substantiate our core claim that architectural nitegration, achieved through specialized pathways and intrinsic mechanisms, surpasses the capabilities of isolated improvements.

\begin{figure*}[t]
    \centering
    \includegraphics[width=0.95\textwidth]{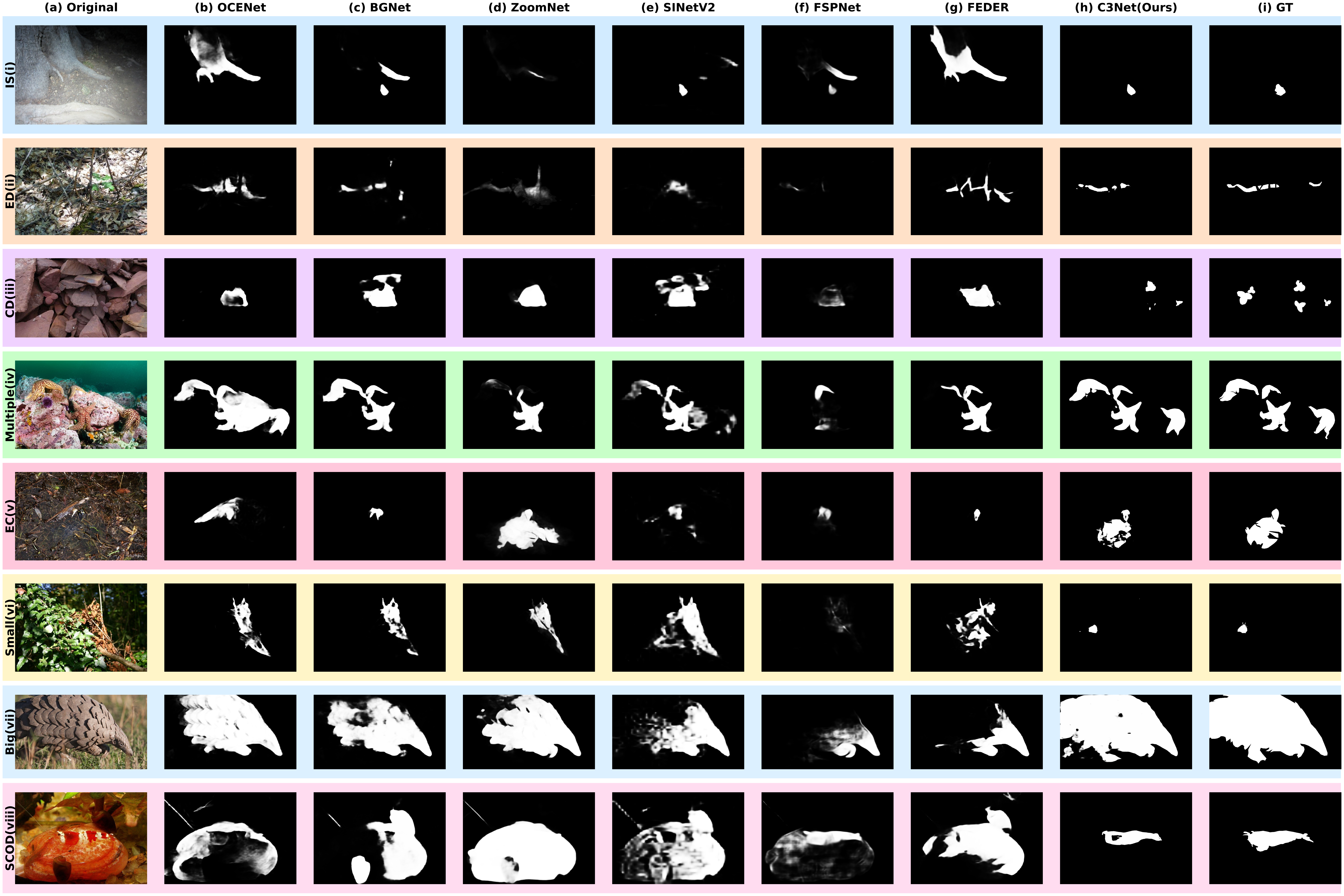}
    \caption{Visual comparison of C3Net with state-of-the-art methods on challenging COD cases. Each row exemplifies a specific challenge: (i) Intrinsic Similarity (IS), (ii) Edge Disruption (ED), (iii) Contextual Dependencies (CD), (iv) Multiple Instances, (v) Environmental Complexities (EC), (vi) Small Objects (ESV aspect), (vii) Large Objects (ESV aspect), and (viii) Salient-Camouflaged Object Disambiguation (SCOD). For each row, columns are: (a) Input Image, (b) OCENet~\cite{9706783}, (c) BGNet~\cite{sun2022bgnet}, (d) ZoomNet~\cite{pang2022zoom}, (e) SINetV2~\cite{fan2021concealed}, (f) FSPNet~\cite{huang2023feature}, (g) FEDER~\cite{He_2023_CVPR}, (h) C3Net (Ours), and (i) Ground Truth. Visual results for the compared methods are obtained from officially released predictions.}
    \label{fig:qualitative}
    \vspace{-3mm}
\end{figure*}

\paragraph{Qualitative Analysis.} Figure~\ref{fig:qualitative} presents visual comparisons that validate our architectural design through consistent superior performance. In IS scenarios (Figure~\ref{fig:qualitative}, row i) where objects share identical textures with backgrounds, C3Net accurately segments the camouflaged target. This success stems from our dual-pathway architecture, which extracts complementary features at multiple scales. ED cases (Figure~\ref{fig:qualitative}, row ii) reveal fragmented boundaries in competing methods while C3Net produces complete masks. Our ERP with gradient-initialized EEMs specifically recovers these disrupted boundaries through specialized edge processing. CD examples (Figure~\ref{fig:qualitative}, row iii) and multiple instance scenarios (Figure~\ref{fig:qualitative}, row iv) demonstrate accurate segmentation where others fail or merge objects. The CLP enables this through deep semantic understanding combined with spatial precision. EC (Figure~\ref{fig:qualitative}, row v) shows C3Net maintaining segmentation integrity under shadows and occlusions while competitors produce fragmented results. ESV is handled robustly with both small objects (Figure~\ref{fig:qualitative}, row vi) and large objects (Figure~\ref{fig:qualitative}, row vii) accurately segmented. Content-adaptive upsampling preserves fine details across these scales. Most significantly, the SCOD challenge (Figure~\ref{fig:qualitative}, row viii) exposes fundamental limitations in existing methods. While OCENet, BGNet, ZoomNet, and FSPNet all erroneously segment the salient orange coral, C3Net correctly identifies only the camouflaged ray. This validates the ICG mechanism's ability to suppress salient distractors through intrinsic context-contrast analysis, without relying on external saliency models. Across all challenges, C3Net consistently produces masks closest to ground truth while competing methods exhibit complete failures or severe artifacts. 

\subsection{Ablation Studies}
\label{subsec:ablation}

We validate C3Net's architectural design through systematic ablation studies across all benchmarks. Table~\ref{tab:ablation} investigates implementation alternatives and component contributions.

\begin{table}[t]
    \centering
    \caption{Ablation study validating architectural components and implementation choices across all benchmarks. Notes: $\uparrow$ denotes that higher values are better, while $\downarrow$ denotes that lower values are better. We evaluate alternative implementations (rows 2-3) by replacing key components with standard alternatives. We then remove essential components (rows 4-7) to analyze their individual contributions. Bold values indicate full C3Net baseline performance.}
    \label{tab:ablation}
    \resizebox{\columnwidth}{!}{
    \begin{tabular}{l|ccc|ccc|ccc}
        \toprule
        \multirow{2}{*}{Configuration}
        & \multicolumn{3}{c|}{CAMO (250)}
        & \multicolumn{3}{c|}{COD10K (2,026)}
        & \multicolumn{3}{c}{NC4K (4,121)} \\
        \cline{2-10}
        & $S_\alpha\uparrow$ & $F_\beta^w\uparrow$ & $\mathcal{M}\downarrow$
        & $S_\alpha\uparrow$ & $F_\beta^w\uparrow$ & $\mathcal{M}\downarrow$
        & $S_\alpha\uparrow$ & $F_\beta^w\uparrow$ & $\mathcal{M}\downarrow$ \\
        \midrule
        C3Net (Full)
        & \textbf{0.904} & \textbf{0.889} & \textbf{0.0311}
        & \textbf{0.898} & \textbf{0.851} & \textbf{0.0162}
        & \textbf{0.913} & \textbf{0.895} & \textbf{0.0220} \\
        \midrule
        w/ ViT-L (no registers)
        & 0.875 & 0.843 & 0.0412
        & 0.871 & 0.806 & 0.0228
        & 0.892 & 0.856 & 0.0278 \\
        w/ Bilinear upsampling
        & 0.889 & 0.871 & 0.0337
        & 0.880 & 0.832 & 0.0182
        & 0.902 & 0.878 & 0.0239 \\
        \midrule
        w/o ERP
        & 0.881 & 0.862 & 0.0341
        & 0.872 & 0.825 & 0.0215
        & 0.897 & 0.871 & 0.0245 \\
        w/o CLP
        & 0.631 & 0.538 & 0.0847
        & 0.608 & 0.520 & 0.0521
        & 0.664 & 0.558 & 0.0598 \\
        w/o ICG mechanism
        & 0.883 & 0.870 & 0.0362
        & 0.875 & 0.829 & 0.0187
        & 0.900 & 0.876 & 0.0257 \\
        w/ Random EEM init
        & 0.893 & 0.876 & 0.0329
        & 0.885 & 0.838 & 0.0186
        & 0.902 & 0.880 & 0.0238 \\
        \bottomrule
    \end{tabular}}
    \vspace{-3mm}
\end{table}

\paragraph{Implementation Choices.} We first examine encoder and upsampling alternatives. Replacing DINOv2 with standard ViT-Large decreases $F_\beta^w$ by 5.3\% on COD10K, 5.2\% on CAMO, and 4.4\% on NC4K. Similarly, replacing DySample with bilinear upsampling causes $F_\beta^w$ reductions of 2.2\% on COD10K, 2.0\% on CAMO, and 1.9\% on NC4K. These moderate degradations show that such choices optimize performance but do not drive core detection capability.

\paragraph{Dual-Pathway Architecture.} Next, we analyze the architectural pathways. Removing either pathway highlights their unique contributions and asymmetric significance. Without ERP, $F_\beta^w$ decreases by 3.1\% on COD10K, 3.0\% on CAMO, and 2.7\% on NC4K. These moderate losses confirm that ERP adds valuable boundary refinement. In sharp contrast, removing CLP causes a severe failure. On COD10K, $S_\alpha$ drops to 0.608 and $F_\beta^w$ falls to 0.520, a 38.9\% decrease in weighted F-measure. CAMO and NC4K experience similar drops, with $F_\beta^w$ values of 0.538 and 0.558. The MAE rises by 221.6\% on COD10K, 172.3\% on CAMO, and 171.8\% on NC4K. CLP has two parts that jointly enhance detection abilities. The SEU blocks provide semantic insight through deep feature processing, while ICG suppresses saliency to counter SCOD. Removing the entire CLP eliminates both functions simultaneously, resulting in a 39\% collapse. This is a 13-fold difference compared to removing ERP. This shows that semantic understanding is far more critical than boundary refinement alone. These results support our dual-pathway design with its clear architectural hierarchy.


\paragraph{Saliency Suppression.} Within CLP, we isolate ICG's specific contribution to saliency handling. The ICG mechanism addresses SCOD by suppressing intrinsic saliency without the need for external models. Without ICG, $F_\beta^w$ decreases by 2.6\% on COD10K, 2.1\% on CAMO, and 2.1\% on NC4K. The MAE increases substantially by 15.4\% on COD10K, 16.4\% on CAMO, and 16.8\% on NC4K. The MAE increases are particularly significant because SCOD failures manifes  t as false positives from salient distractors. These impacts demonstrate that ICG provides a crucial capability for handling saliency. This complements the semantic understanding from SEU blocks to achieve comprehensive detection performance.

\paragraph{Edge Initialization.} Eventually, we investigate the initialization strategy for the ERP components. Our gradient-based initialization strategy consistently improves performance. Random EEM initialization causes $F_\beta^w$ drops of 1.5\% on COD10K, 1.4\% on CAMO, and 1.6\% on NC4K. These results show that classical edge detection priors provide useful inductive biases for boundary detection.

These results establish clear architectural priorities. CLP provides foundational semantic understanding through its SEU blocks. ICG contributes crucial saliency suppression to address SCOD challenges. ERP provides valuable boundary refinement. Together, these architectural components enable core detection. In contrast, implementation choices such as encoder selection and upsampling method offer meaningful optimizations but are not essential for detection. Similarly, initialization strategy offers minor improvements through inductive biases. The 13-fold difference between CLP and ERP removal confirms that architectural design choices matter far more than implementation alternatives.
\section{Conclusion}
\label{sec:conclusion}

We presented C3Net for comprehensive camouflaged object detection, addressing six fundamental challenges. Our dual-pathway architecture achieves state-of-the-art performance through specialized processing. The ERP with gradient-initialized EEMs captures precise boundaries, while the CLP with ICG suppresses salient distractors intrinsically without external models. Our architecture outperforms previous state-of-the-art methods while maintaining efficient processing. The key innovations—dual-pathway separation, intrinsic saliency suppression, gradient-initialized edge detection, and attentive fusion—enable critical applications in medical imaging and wildlife monitoring. Our C3Net demonstrates that complex vision challenges require specialized components working synergistically rather than isolated improvements.


\bibliographystyle{IEEEtran}
\bibliography{references}

\end{document}